# Electromagnetic Scattering Kernel Guided Reciprocal Point Learning for SAR Open-Set Recognition

Xiayang Xiao†, Zhuoxuan Li†, *Student Member*, Ruyi Zhang, Jiacheng Chen, Haipeng Wang, Senior *Member, IEEE*

*Abstract*—The limitations of existing Synthetic Aperture Radar (SAR) Automatic Target Recognition (ATR) methods lie in their confinement by the closed-environment assumption, hindering their effective and robust handling of unknown target categories in open environments. Open Set Recognition (OSR), a pivotal facet for algorithmic practicality, intends to categorize known classes while denoting unknown ones as "unknown." The chief challenge in OSR is to mitigate risks associated with numerous unknown samples within the open space. To enhance open-set SAR classification, a method called scattering kernel with reciprocal learning network is proposed. Initially, a feature representing paradigm is constructed on the base of reciprocal point learning (RPL), establishing a bounded space for potential unknown classes. This approach indirectly introduces unknown information into a learner confined to known classes, thereby acquiring more concise and discriminative representations. Subsequently, considering the variability in the imaging of targets at different angles and the discreteness of components in SAR images, a proposal is made to design large-sized convolutional kernels based on attribute scattering center(ASC) models. This enhances the ability to extract intrinsic non-linear features and specific scattering characteristics in SAR images, thereby improving the discriminative features of the model and mitigating the impact of imaging variations on classification performance. Experiments conducted on the public datasets substantiate the superior performance of this approach called ASC-RPL over mainstream methods.

## I. INTRODUCTION

As we know, the convolutional neural network (CNN) technology has significantly propelled the development of SAR remote sensing image analysis, particularly playing a pivotal role in target detection and recognition [1]-[5]. Inspired by the cognitive mechanisms of the human brain, CNNs construct a multi-layered feature extraction structure through multiple non-linear transformations, enabling the nonlinear mapping of raw data into a multidimensional feature domain conducive to accurately classifying targets [6]. Typically, a CNN model employs a softmax classifier [7] at its end to output the confidence in target classification, evaluated under closed-set conditions. Under the closed-world assumption, applying softmax operations during training necessitates the awareness of all test categories, which subsequently fails to retain probabilities for any unknown classes.

However, the real-world environment is continually changing and open, imposing various objective limitations on classifiers, making it impractical during the training phase to exhaustively test all possible categories. During testing, unknown categories may be submitted to the algorithmic model, violating the closed-world assumption of recognition methods, rendering the original recognition system ineffective [8]. Typical CNNs use linear classification layers and softmax on embedded features to generate probability distributions for known categories, thereby assuming samples from unknown categories have a uniform distribution across known categories. This leads to softmax loss solely increasing feature separability without effectively distinguishing between known and unknown categories. Specifically, the model might yield high confidence scores for unknown categories but actually misclassify them. Therefore, outstanding performance under closed-set settings is illusory and does not meet the requirements of real-world applications. A robust recognition system should be capable of distinguishing test samples into known and unknown categories and correctly classifying known categories. This task is commonly referred to as Open Set Recognition (OSR) in more realistic scenarios [9].

This paper focuses on addressing the identification issues of SAR images in open environments, namely SAR Open Set Recognition (SOSR). The aim is to boost the CNN's capability to reject unknown categories while sustaining its high recognition performance for known categories. In the context of Open Set Recognition, CNNs face the following challenges posed by SAR images: 1) Detectors need to mitigate the classification risk of known data under strong supervision, minimizing the overlap between known data distribution in feature space and potential distribution of unknown data to reject unknown samples. 2) Existing deep learning-based SAR image classification networks primarily rely on fine-tuning optical networks, without taking into account the unique imaging mechanism and scattering characteristics of SAR images. Additionally, the underlying physical decision logic of the original CNN networks remains unclear. Unlike optical satellite imaging mechanisms, SAR is sensitive to imaging conditions and observation angles. Directly transferring

This work was supported in part by the National Natural Science Foundation of China (Grant No. 62271153) and the Natural Science Foundation of Shanghai (Grant No. 22ZR1406700). (Corresponding author: Haipeng Wang, e-mail: hpwang@fudan.edu.cn)
Xiayang Xiao is with the Key Laboratory for Information Science of Electromagnetic Waves (Ministry of Education), School of Information Science and Technology, Fudan University, Shanghai 200433, China, and also with China Mobile Internet Company Limited, Guangzhou 510030, China.
Zhuoxuan Li, Ruyi Zhang,Jiacheng Chen, Haipeng Wang are with the Key Laboratory for Information Science of Electromagnetic Waves (Ministry of Education), School of Information Science and Technology, Fudan University, Shanghai 200433, China.
† These authors contributed equally to this work

optical networks to SAR images fails to achieve optimal results. Therefore, the key issue lies in how to fully leverage the scattering characteristics of SAR images to enhance the interpretability and decision transparency of the network, ultimately achieving a more robust performance.

In SAR open-set recognition, a pivotal challenge lies in mitigating the risk of misclassifying potential unknown classes as known classes. To address this challenge, prototype learning [9]-[12] is employed to represent known classes in the embedding space and encourage the characteristics of the training set classes to converge towards their corresponding prototypes. However, the learned prototype representations may gradually approach the space of unknown classes during training, leading to an overlap in their feature distributions. This occurs because these methods only consider data from known classes, disregarding potential features of unknown data, which poses significant risk in open spaces. Drawing upon the aforementioned analysis, this study proposes the utilization of reciprocal points to model the latent unknown deep space, thereby mitigating the risks inherent in open spatial domains. Reciprocal points, construed as inverse prototypes of known categories, can be comprehended as instantiated representations of the latent out-of-class space [13]. During the training phase, all samples from known classes are pushed to the periphery of the space for classification with the assistance of their corresponding reciprocal points and constrained by boundary limits. The unknown category, along with the reciprocal points, forms the outer space of categories, with the reciprocal point set restricting its range, and bounded constraints preventing the algorithm from assigning excessively high confidence to unknown categories. By leveraging reciprocal points to minimize the overlap in feature distribution between known and unknown classes, unknown categories become more easily distinguishable. This method provides better distinctiveness for open set recognition tasks, improving system performance when faced with unknown categories and offering new possibilities.

Furthermore, a study [14] indicates a significant correlation between closed-set and open-set performance. The study confirms the crucial importance of enhancing recognition accuracy within the closed-set to effectively optimize open-set recognition outcomes. It repositions the OSR problem as part of traditional recognition problems. Therefore, robust feature representation is crucial for enhancing a model's performance in open-set tasks. Current mainstream SAR classification methods primarily rely on networks like CNNs and Transformers. However, most studies utilize network structures based on optical images, disregarding SAR targets' intrinsic characteristics, limiting improvements in recognition accuracy and generalization capabilities. Li et al. [15] pointed out that the primary challenge in SAR target detection lies in the domain transfer from RGB to SAR, as the visual characteristics of the images differ significantly. In contrast to deep learning approaches, traditional SAR target classification algorithms provide interpretable SAR target features. Common SAR target features encompass geometric features [16], transformation domain features [17], and electromagnetic scattering features (ESF)[18]. Target classification algorithms based on ESF are currently receiving considerable attention. Within high-frequency regions, the radar cross-section of extended targets is amenable to approximation as the aggregate of the scattering contributions from discrete centers. The Attribute Scattering Center (ASC) model [19] effectively describes the dependency of the target's radar backscatter on both frequency and the azimuthal angle. This model includes abundant physical and geometric attributes, providing a more accurate description of radar targets' electromagnetic scattering characteristics in high-frequency areas. Consequently, the ASC model effectively reflects the physical features of SAR targets. Considering the unique attributes of SAR images in the scattering domain[20], this study proposes the design of a large-sized convolutional kernel based on attribute scattering centers, enabling the network to better extract scattering features from SAR images, thereby enhancing the accuracy and robustness of the model.

Based on the aforementioned, this study introduces an open-set recognition network structure integrating ASC characteristics within convolutional kernels to enhance SAR classification performance under open-set conditions. As illustrated in Fig. 1, this framework retains the representation capability of CNN while discarding the closed-world assumption of softmax, employing a reciprocal point-based model to classify known targets and discern unknown targets. Specifically, to enhance the network's representation of SAR target features, this work firstly constructs an ASC-based network architecture. Parameters in the convolutional kernels are modulated using ASC that reflects target scattering characteristics, allowing the network to better focus on the scattering features of SAR targets. Furthermore, for varying sample sizes in classification scenarios, a larger convolutional kernel structure was designed to further enhance recognition performance. Finally, within the classification head, reciprocal point learning is designed for SAR image object open-set recognition, addressing issues of modeling unknown data and mitigating risks within open spaces. Additionally, a bounded constraint is devised to avert the neural network from assigning excessively high confidence scores to unidentified samples, thereby improving the separability between known and unknown categories. Ultimately, extensive experiments conducted on the MSTAR dataset validated the effectiveness of the proposed methodology.

The contributions can be summarized as follows:

1 ) The designed large-sized ASC kernel, in comparison to conventional convolutional kernels, offers a larger receptive field. It demonstrates higher focus and efficiency in extracting certain intrinsic non-linear features and specific scattering characteristics in SAR images.

2 ) Addressing the modeling of out-of-class space through known category-based risk reduction in open spatial domains, reciprocal point learning is proposed to address SAR image target open-set recognition. This approach tackles the challenge of unknown data modeling, thereby reducing the risks associated with open space. As far as we are aware, this study is the first application of reciprocal point in SAR object recognition.

3) The original CNN network lacks explicit physical decision logic. The designed model in this paper elucidates the physical logic behind the network's discriminative decisions, enhancing interpretability and decision transparency. This augmentation contributes to the algorithm model's robustness. Concurrently, visualization results substantiate the complementary relationship between deep abstract features and electromagnetic scattering features.

## II. PRELIMINARY KNOWLEDGE

### A. Paradigms and Limitations of Deep Learning Models

Traditional CNNs are generally divided into two main parts [21]: the feature extraction function $f(x,\theta)$ and classifier $C(f(x,\theta),\theta_c)$. It can achieve the traditional object recognition task by learning the mapping function $f: X \to Y$, mapping the input image domain X to the corresponding label domain Y. This mapping function model dictates the network's performance and efficiency in handling various image data. To better adapt the network to the relationship between samples and supervision information, optimization of trainable parameters is typically necessary. The core of this optimization process involves adjusting the model's parameters by minimizing a loss function to better fit the training data.

Formula (1) illustrates that in the classifier $C(f(x,\theta),\theta_c)$, it's usually assumed during training that the sum of the posterior probabilities of known classes is 1, namely:

$$p(y|x) = \frac{e^{(w_y \cdot f(x;\theta)+b_y)}}{\sum_{i=1}^{C} e^{(w_i \cdot f(x;\theta)+b_i)}} \quad (1)$$

$$\sum_{i=1}^{C} p(i|x) = 1 \quad (2)$$

where $y$ represents the actual label of sample $x$. The $w_i$ and $b_i$ represent the weight vector and bias parameters for the $i$-th class, respectively. It's worth noting that this framework exclusively considers training for known classes and doesn't retain probability information for unknown classes, thus becoming ineffective when dealing with unknown categories.

Formula (3) below represents the traditional training loss function:

$$loss = \min_{f \in F} \sum_{i=1}^{N} \ell(f(x_i;\theta), y_i) \quad (3)$$

where $x_i$ and $y_i$ denote the i-th sample and its corresponding label, $N$ represents the total number of targets, and $\ell$ denotes the network's objective loss function.

Considering the physical model $g: X \to Z$ providing additional prior information by mapping samples X to a physical representation $Z$, which incorporates a deep neural network of $Z$, learns to fit the joint distribution $f(X,Z)$ of images and physical priors [22]. Therefore, its minimized objective function can be expressed as:

$$loss = \min_{f \in F} \sum_{i=1}^{N} \ell(f(x_i;\theta;g(x_i)), y_i) \quad (4)$$

where $z_i = g(x_i)$ denotes the prior information of the $i$-th sample $x_i$. To effectively learn discriminative features of SAR targets, networks typically require training on samples from the same or similar distributions. However, when observation conditions change, generalization performance might be affected, making robust prior knowledge crucial for addressing automatic recognition problems caused by differences in image modalities.

### B. Open-set Recognition for SAR Images

In the domain of SAR image analysis, the open-set problem on the MSTAR dataset was tackled by Scherreik et al. [23] using Weibull-calibrated SVM (W-SVM) [24] and Probabilistic Open Set SVM (POSSVM) [25]. Zeng et al. [26] employed the Kullback-Leibler Divergence (KLD) between test and training set features to roughly identify unknown SAR targets. Generative Adversarial Networks (GAN) [27] were utilized for SAR image open-set recognition, distinguishing between unknown and known targets based on a branch output score compared against a predefined threshold. Inkawhich et al. [28] proposed Adversarial Out-of-Distribution Exposure (AdvOE) to jointly design SAR automatic detection systems, ensuring accuracy and Out-of-Distribution (OOD) detection. Dang et al. [29] introduced a classifier for continuously detecting and learning new categories in an incremental learning scenario, leveraging the MSTAR dataset. Subsequently, in [30], they proposed an extended method to retain prior recognition capabilities when adding new tasks/categories. These studies addressed SAR's OSR problem through threshold-based decisions, reconstruction error-based judgments, and incremental learning. However, the intrinsic characteristics of SAR targets were overlooked in their processing, limiting the improvement of identification precision and weaknesses in generalization capabilities.

## III. METHODOLOGY

### A. Kernel Based on ASC

The Attributed Scattering Center Model (ASC) model is developed from the principles of physical optics and the geometric theory of diffraction, with its parameters reflecting a variety of physical and geometric characteristics that describe the electromagnetic properties of typical ASCs. In the high-frequency domain, as shown by the following equation, the radar echo from targets can be considered as the superposition of multiple ASCs:

$$E(f,\varphi;\Theta) = \sum_{i=1}^{q} E_i(f,\varphi;\theta_i) + n(f,\varphi;\theta_i) \quad (5)$$

where $f$ denotes the operating frequency in Hertz (Hz); $\varphi$ represents azimuthal angle, $\Theta$ encompasses the parameter set of the scattering center. $n(f, \varphi; \theta_i)$ signifies noise, and the total number of ASCs within the target is defined as $q$.

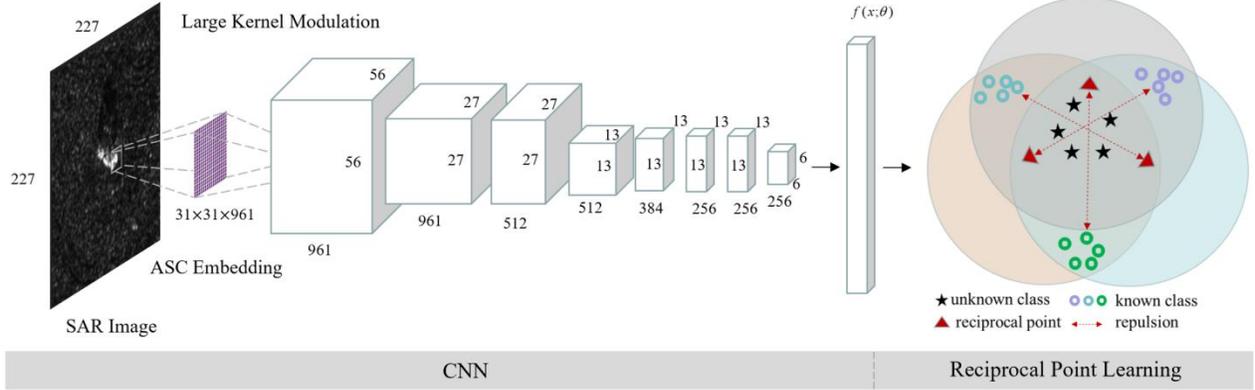

**Fig. 1** The overall architecture of the proposed method

The scattering field of distributed targets in the high-frequency domain can be represented as follows:

$$E_i(f, \varphi; \theta_i) = A_i \cdot (j\frac{f}{f_c})^{\alpha_i} \cdot \exp[-j\frac{4\pi f}{c}(x_i\cos\varphi + y_i\sin\varphi)]$$
$$\cdot \text{sinc}(\frac{2\pi f}{c}L_i\sin(\varphi - \overline{\varphi}_i)) \cdot \exp(-2\pi f\gamma_i\sin\varphi) \quad (6)$$

where $sinc(x) = (\sin(\pi x)/(\pi x))$, $f_c$ is the radar center frequency and $c$ means is the light propagation. $(x_i, y_i)$ denote the coordinates of the scattering center $i$ in the cross-range and range direction, respectively. $A_i$ is the relative amplitude, $L_i$ and $\overline{\varphi}_i$ represent the length and pointing angle of the distributed of the scattering center $i$, respectively. The parameter $\alpha_i$ presents requency-dependent factor, and $\gamma_i$ is the aspect dependence. $\theta_i = [A_i, \alpha_i, x_i, y_i, L_i, \overline{\varphi}_i, \gamma_i]$, represents the parameters associated with the i-th scattering center. In the case of local scattering centers, $L_i = \overline{\varphi}_i = 0$. whereas for distributed scattering centers, $\gamma_i = 0$.

Taking into account the actual working conditions of a SAR system, certain approximations can be applied to the ASC model [31]: The angle observation is generally short, permitting the angle dependence factor $\gamma_i$ to be effectively neglected. Additionally, the ratio of center frequency to bandwidth is generally low, meaning that the parameter $\alpha_i$ has minimal impact. As a result, in this study, $\gamma_i$ and $\alpha_i$ are set to zero. With these assumptions, the ASC can be simplified as follows:

$$E_i(f, \varphi; \theta_i) \approx A_i \text{sinc}[2\frac{f}{c}L_i\sin(\varphi - \overline{\varphi}_i)]$$
$$\times \exp[-j4\pi\frac{f}{c}(x_i\cos\varphi + y_i\sin\varphi)] \quad (7)$$

As shown in Fig. 2, the convolution process and its results are clearly depicted. In this procedure, the convolution operation emphasizes the primary structural features of the original image, predominantly encompassing shape and position, correlated with the characteristics of the convolution kernel. This phenomenon reveals the crucial role of convolution in feature extraction and its substantial impact on image analysis. Based on the fundamental principles of convolution kernels and the ASC model, we designed ASC kernels with diverse parameters. These ASC kernels exhibit higher efficiency compared to traditional convolutional kernels in extracting certain intrinsic nonlinear features and specific scattering characteristics in SAR images.

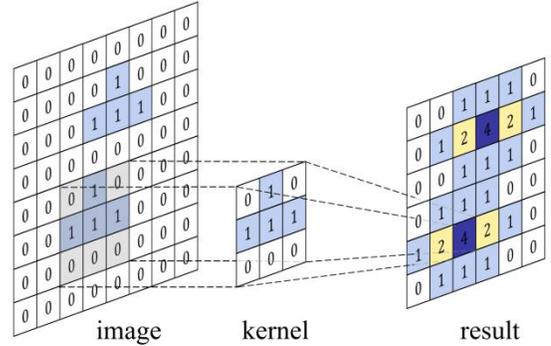

Fig. 2 The convolution process and results. the convolution operation emphasizes the primary structural features of the original image, predominantly encompassing shape and position, correlated with the characteristics of the convolution kernel.

To obtain the ASC-modulated convolution kernels, it is essential to pre-generate a set of ASC kernels. As per Equation (5), each simplified ASC comprises five attribute parameters denoted as $\theta = [A, x, y, L, \overline{\varphi}]$. To generate the ASC kernels, each position of ASC at the kernel center, setting to zero. Subsequently, the amplitude of the scattering center is normalized as 1. Equation (7) is then transformed into Equation (8).

$$E_i(f, \varphi; \theta_i) \approx \text{sinc}[2\frac{f}{c}L_i\sin(\varphi - \overline{\varphi}_i)] \quad (8)$$

In this study, $\overline{\varphi}$ is uniformly sampled between certain values, L is governed by both the spatial resolution of the image and the size of the CNN's convolution kernel, where the spatial resolution corresponds to the physical distance between adjacent pixels in the input image. Assuming a spatial resolution of m, and a kernel size of $r \times r$, L can be calculated using Equation (9). The specific weights of the

ASC kernels can be computed based on the radar imaging parameters of the dataset.

$$L = p \times m \ (p \in \mathbb{N}, p \leq r) \quad (9)$$

Performing a 2D-FFT on Equation (8) yields the corresponding scattering center image, as expressed in Equation (10):

$$ASC_i = \text{FFT}\{E_i(f, \varphi; \theta_i)\} \quad (10)$$

By cropping the scattering center image from the center based on the convolution kernel size, the ASC convolution kernel can be obtained:

$$ASC_{K_i} = \text{CropCenter}(ASC_i, r) \quad (11)$$

Initializing the first layer of the CNN's convolutional kernels with the obtained weights allows the network to extract electromagnetic scattering features from SAR images:

$$CNN_{first\_layer} = \text{Intialize}(ASC_{K_i}) \quad (12)$$

The network architecture used in the experiment is based on a modified version of AlexNet. The framework consists of five convolutional layers, three pooling layers, and three fully connected layers. Unlike AlexNet, we have improved the first convolutional layer by setting it with a large-size ASC convolutional kernel. The detailed settings for each layer are presented in TABLE I.

TABLE I THE CONFIGURATION OF THE NETWORK STRUCTURE-31×31 ASC KERNEL

| Layer | Kernel Size | Kernel num. | Stride | Padding | Output |
|---|---|---|---|---|---|
| Input | - | - | - | - | $227 \times 227 \times 1$ |
| Conv1 | $31 \times 31$ | 961 | $4 \times 4$ | $2 \times 2$ | $56 \times 56 \times 961$ |
| Pool | $3 \times 3$ | - | $2 \times 2$ | - | $27 \times 27 \times 961$ |
| Conv2 | $5 \times 5$ | 512 | $1 \times 1$ | $2 \times 2$ | $27 \times 27 \times 512$ |
| Pool | $3 \times 3$ | - | $2 \times 2$ | - | $13 \times 13 \times 512$ |
| Conv3 | $3 \times 3$ | 384 | $1 \times 1$ | $1 \times 1$ | $13 \times 13 \times 384$ |
| Conv4 | $3 \times 3$ | 256 | $1 \times 1$ | $1 \times 1$ | $13 \times 13 \times 256$ |
| Conv5 | $3 \times 3$ | 256 | $1 \times 1$ | $1 \times 1$ | $13 \times 13 \times 256$ |
| Pool | $3 \times 3$ | - | $2 \times 2$ | - | $6 \times 6 \times 256$ |
| FC1 | $6 \times 6 \times 256$ | | | | 1024 |
| FC2 | 1024 | | | | 1024 |
| FC3 | 1024 | | | | 10 |

*B. Large Kernel Design*

The emergence of Vision Transformers (ViT) [32] has posed new challenges to CNNs, showcasing remarkable performance across various visual tasks and prompting a reevaluation of the effectiveness of CNNs. In Vision Transformers, the design of Multi-Head Self-Attention (MHSA) has proven crucial, whether global [33] or local. The larger kernel in MHSA allows each MHSA layer's output to gather information from a wide region, effectively enhancing the model's receptive field. In contrast to ViT, Convolutional Neural Networks address the receptive field issue differently, favoring the use of a series of small spatial convolutions [34], such as 3×3, rather than widely adopting large kernels. Recent research [35] has attempted to maintain comparable results by introducing 7×7 depth convolutions as an alternative to MHSA layers. While this approach shares similarities with ViT's design, its motivation differs. The study does not extensively explore the relationship between large kernels and performance nor elaborate on the application of large kernels in conventional CNNs. Instead, it attributes the outstanding performance of Vision Transformers to sparse connections, parameter sharing, and dynamic mechanisms, highlighting differences in understanding and emphasis on performance improvement among various research endeavors.

The design of large kernels can effectively increase receptive field, thus ameliorating shape distortions [36]. Although achieving a large receptive field can be realized by combining a series of small kernels, for instance, decomposing a 7×7 convolutional kernel into a stack of three 3×3 kernels to ensure information preservation (albeit requiring additional channels to maintain model flexibility), its effectiveness may not always be as pronounced as using a large convolutional kernel directly in certain scenarios.

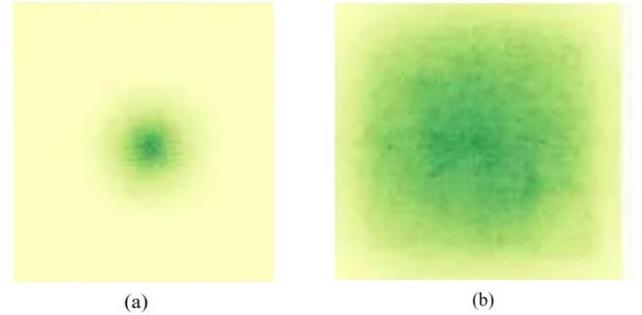

Fig. 3 The effective receptive fields. (a)The receptive field of ResNet101; (b) The receptive field of the RePLKNet model with a large 31x31 convolutional kernel.

A large receptive field is achieved through the design of larger kernels. The Effective Receptive Field (ERF) theory [37] establishes a direct correlation between ERF and the size of kernels, as well as the depth of the CNN layers. This signifies that an increase in kernel size and layer depth enables the network to assimilate a broader spectrum of the input image. However, expanding the depth of convolutional layers introduces optimization challenges [38], complicating network training as depth escalates. Despite the successful training of networks with hundreds of layers in some deep network architectures like Residual Networks (ResNets), research suggests that their actual depth might not be as significant as apparent [39][40]. Reffering to Fig.3, research [40] demonstrates ResNets behaving akin to multiple shallow networks, indicating limited ERF despite substantial depth increments. This observation aligns with previous empirical studies [41]. In essence, larger kernel designs offer advantages in achieving a substantial ERF while circumventing optimization issues from increased depth.

The principles of receptive fields still adhere to general guidelines when developing deep learning models based on SAR images. Therefore, this study explores the possibility of optimizing traditional network structures by designing large-sized convolutional kernels within the context of the ASC convolutional kernel model.

Therefore, this chapter's experiments involved improving the initial convolutional kernel of the convolutional neural network using $31 \times 31$, $21 \times 21$, and $11 \times 11$ kernel sizes, respectively, to broaden the receptive field. Correspondingly, the number of convolutional kernels in the first layer are 100,441,961. TABLE I presents an example configuration of the layers using the 31x31 convolutional kernels. The analysis of ablation results are presented in TABLE VI, affirming the potential of larger kernel designs in enhancing convolutional neural network performance.

*C. Reciprocal Point Learning*

The reciprocal points are regarded as the inverse prototypes of each known category[13]. In this study, it is designed to establish a bounded space for potential unknown classes.. During training, reciprocal point learning increases the distance in feature space between known categories and its reciprocal points by pushing all known categories toward the outer regions of the space, helping to separate them from unknown categories.

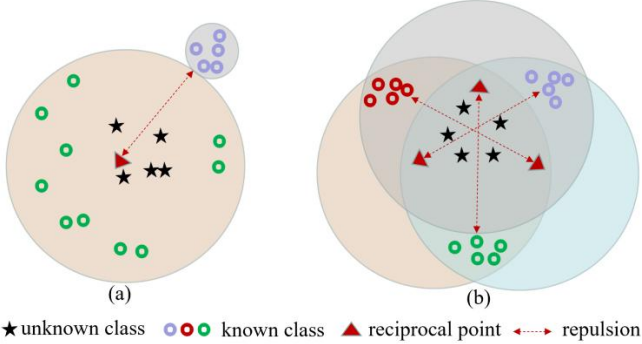

★ unknown class   ○○○ known class   ▲ reciprocal point   ←---→ repulsion

Fig. 4 The depiction of RPL. (a) For a specific known category; (b) The explanation of the relations between known categories and the corresponding reciprocal points.

Specifically, given a set of training sample $D_L = \{(x_1, y_1), ..., (x_n, y_n)\}$, it comprises $N$ known classes denoted as $C_k = \{1, ..., N\}$. The test dataset: $D_T = \{t_1, ..., t_u\}$, $t_i \in C_k \cup C_u$, $C_u = \{N = N + 1, ..., N + u\}$, and $u$ represents the quantity of unseen categories in the actual testing environment. The potential unknown data is represented as $D_u$. Known classes within the $m$-dimensional feature space $R_m$ are denoted as $S_k$, while the space of unknown classes can be expressed as $O_k = R_m - S_k$.

Therefore, the reciprocal point $p^k$ of category k can be considered as the a probable depiction of the subset of data $D_L^{\neq k} \cup D_u$. In Fig.4(a), contrary to prototype learning, the reciprocal point $P^k$ should be more similar to the samples in $O_k$. It is closer to $O_k$ in the feature space and further away from $S_k$. It can be expressed as:

$$\forall d \in \zeta(D_L^k, P^k), \max(\zeta(D_L^{\neq k} \cup D_u, P^k)) \leq d \quad (12)$$

$\zeta()$ represents the calculation of the distance set of all samples between the two groups; $max()$ is the maximum value function. The goal is to separate unknown categories from known categories as much as possible in the feature space. Based on this, the classification of samples can be facilitated by augmenting the separation between the reciprocal point and the associated known class. The distance between the known category $k$ and $P^k$ is the largest. Specifically, the reciprocal points $P^k$ of known categories are optimized through the learnable feature representation network $f_\theta(x)$. Given that the spatial orientation and angular alignment of each recognized category are in direct opposition to their respective reciprocal points, the Euclidean distance $d_e$ and dot product $d_d$ are used to evaluate the similarity:

$$d_e(f_\theta(x), P^k) = \frac{1}{m} \cdot \left\| f_\theta(x) - P^k \right\|_2^2 \quad (13)$$

$$d_d(f_\theta(x), P^k) = f_\theta(x) \cdot P^k \quad (14)$$

$$d(f_\theta(x), P^k) = d_e(f_\theta(x), P^k) - (f_\theta(x), P^k) \quad (15)$$

Based on distance measurement, the assignment of $x$ to a specific class can be determined by assessing the discrepancy between the embedded feature $f_\theta(x)$ and the reciprocal points of each class. In contrast to prototype learning, the characteristics of reciprocal points dictate that a larger $d(f_\theta(x), P^k)$ value results in a higher probability of $x$ being assigned to class $k$. The ultimate probability representation can be derived through softmax.

$$p(y = k | x, f_\theta, P) = \frac{e^{\gamma d(f_\theta(x), P^k)}}{\sum_{i=1}^{N} e^{\gamma d(f_\theta(x), P^i)}} \quad (16)$$

The loss function defines the negative log probability of category $k$ as:

$$L_C(x; \theta, P) = -\log p(y = k | x, f_\theta, P) \quad (17)$$

The optimization of the learnable parameters $\theta$ in the feature representation network $f_\theta(x)$ is achieved by minimizing the loss function $L_C$. This optimization aims to enhance the embedding function's ability to classify known categories. Simultaneously, effective separation between known and unknown categories in the feature space is achieved by maximizing the distances between the mutual points and their corresponding samples. As illustrated in Fig.4 (b), during the training phase, each recognized category is propelled towards the periphery of the feature space by their respective mutual points, while the mutual points outside these known categories are pushed away, creating separation from the known space.

However, potential unknown categories are also encompassed within the space outside the class, thereby maintaining the risk associated with open space. Therefore, it is necessary to constrain unknown categories to a bounded unknown space through the utilization of the reciprocal points, as expressed by Equation (18):

$$\max(\zeta(D_L^{\neq k} \cup D_u, P^k)) \leq R \quad (18)$$

where the learnable boundary is denoted as $R$ constraining potential unknown category boundaries. Since the spaces $S_k$ and $O_k$ are complementary, the distance between the samples in $S_k$ and the reciprocal point $P^k$ can be constrained within R by optimizing the margin loss function, thereby indirectly limiting the risk of open space. The boundary loss function

can be expressed as:

$$L_O(x; \theta, P, R) = \max(d_e(f_\theta(x), P^k) - R, 0) \quad (19)$$

The overall loss function for reciprocal point learning can be described as follows:

$$L(x; \theta, P, R) = L_C(x; \theta, P) + \lambda L_O(x; \theta, P, R) \quad (20)$$

where $\lambda$ is a hyperparameter that controls the weight of the open space risk reduction module. $\theta, P, R$ represent the learnable parameters of the framework.

| Algorithm 1: The Training and Testing of RPL for Open-set SAR Recognition | |
|---|---|
| Input: | The SAR image $x_i$ with label $y_i$ for training <br> The convolution network $f_\theta(x)$ with initial parameters $\theta$; The reciprocal points $P$ and the constrained boundary $R$ <br> Hyper-parameter: $\lambda, \gamma$ |
| Output: | The Optimized Parameters, $\theta, P, R$ |
| 1: | **When** $f_\theta(x)$ is not converge **do** |
| 2: | Calculate the loss by $L = L_C + \lambda L_O$ |
| 3: | Update the Parameters $\theta, P, R$ by the back propagation |
| 4: | **end while** |
| Input: | SAR image $x$ |
| Output: | The predicted open-set label $\hat{y}$ |
| 1: | Calculate the distance d between $x$ and $P$ |
| 2: | Predict the closet-set label $\hat{y}$ |
| 3: | **If** $d(\hat{y}) < R$ **then** |
| 4: | $\hat{y} = unknow$ |
| 5: | **end if** |
| 6: | **return** $\hat{y}$ |

## IV. EXPERIMENTS AND ANALYSIS

### A. Datasets Preparation and Evaluation Indicators

The MSTAR dataset collected by Sandia National Laboratories' SAR sensor platform is selected as the baseline for the study [42]. This SAR sensor operates in the X-band and uses HH polarization and the data set is widely used in SAR automatic target recognition research. Each data point is a complex image that can be decomposed into amplitude and phase. The image has a resolution of 0.3 meters × 0.3 meters and a size of 128 × 128 pixels, covering 360° in all directions. These targets can be observed in Fig.5.

TABLE II TRAINING IMAGES AND TEST IMAGES UNDER SOC CONDITION

| Categories | Series Number | Trainset | | Testset | |
|---|---|---|---|---|---|
| | | Depression | Number | Depression | Number |
| BMP2 | 9563 | 17° | 233 | 15° | 195 |
| BTR70 | c71 | 17° | 233 | 15° | 196 |
| T72 | 132 | 17° | 232 | 15° | 196 |
| T62 | A51 | 17° | 299 | 15° | 273 |
| BRDM2 | E71 | 17° | 298 | 15° | 274 |
| BTR60 | 7532 | 17° | 256 | 15° | 195 |
| ZSU23/4 | d08 | 17° | 299 | 15° | 274 |
| D7 | 13015 | 17° | 299 | 15° | 274 |
| ZIL131 | E12 | 17° | 299 | 15° | 274 |
| 2S1 | B01 | 17° | 299 | 15° | 274 |

Different operating conditions (OC) result in varying image distributions. Standard Operating Conditions (SOC) refer to acquiring targets in similar or even identical radar imaging conditions. Besides, Extended Operating Conditions (EOC) signify significant differences between testing and training image acquisitions, including variations in elevation angle, noise interference, and configuration variables. There are 1747 training images and 2425 testing images used. Specific details about the MSTAR dataset are provided in TABLE II.

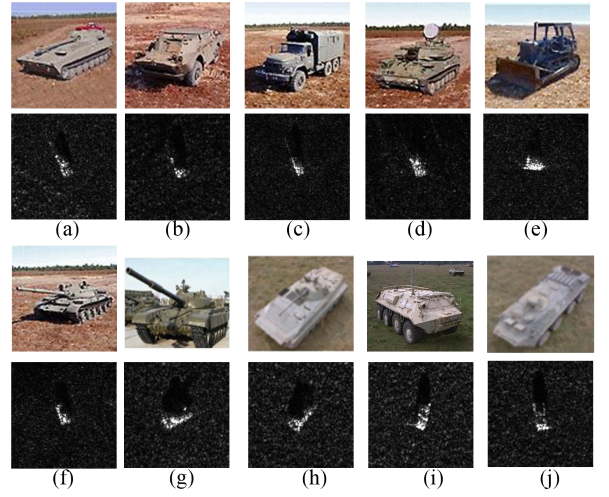

Fig. 5 The optics of the target in MSTAR and its corresponding SAR image. (a) 2S1 (b) BRDM (c) ZIL131 (d) ZSU234 (e) D7 (f) T62 (g) T72 (h) BMP2 (i) BTR60 (j) BTR70

To assess this study, the experimental results are evaluated using the following metrics: Recall, Precision, F1, and Accuracy. Their formulas are as follows:

$$Recall = \frac{\sum_{i=1}^{C} recall_i}{C}, recall_i = \frac{TP_i}{TP_i + FN_i} \quad (21)$$

$$Precision = \frac{\sum_{i=1}^{C} precision_i}{C},$$
$$precision_i = \frac{TP_i}{TP_i + FP_i} \quad (22)$$

$$F1 = \frac{\sum_{i=1}^{C} F1_i}{C}, F1_i = \frac{2 \times precision_i \times recall_i}{precision_i + recall_i} \quad (23)$$

$$Accuracy = \frac{\sum_{i=1}^{C}(TP_i + TN_i) + TU}{\sum_{i=1}^{C}(TP_i + TN_i + TP_i + TN_i) + + (FU + TU)} \quad (24)$$

Additionally, $TU$ and $FU$ represent the correct and incorrect rejections of unknown classes. Furthermore, in order to investigate the robustness of this method in scenarios with varying proportions of known and unknown category, the openness [9] is specifically defined as follows:

$$Openness = 1 - \sqrt{\frac{2 \times |C_{TR}|}{|C_{TR}| + |C_{TE}|}} \quad (25)$$

$C_{TR}$ and $C_{TE}$ represent the number category of train set and test set, respectively. An openness value closer to 1 indicates a higher number of unknown classes in the testset, whereas an openness value of 0 signifies a problem equivalent to closed-set classification.

## B. Implementation Details

In the MSTAR dataset, there are not only complex images used for training and testing, but also comprehensive details about the imaging process, including a carrier frequency $f_c$ of 9.6GHz, and the bandwidth B of 0.49GHz. The frequency (f) range in the attribute scattering center model is set within the interval of 9.36GHz and 9.85GHz. The sampling frequency is 0.591GHz, resulting in approximately 189 samples for f, adding 19 zeros at both ends. $\Delta\varphi$ is 2.8, so the range of $\varphi$ is from -1.4 to 1.4. There are 227 sampling points, matching the size of the input SAR image. For a convolutional kernel size of 11 × 11, $L_i$ is set to {0.3, 0.6, 0.9, 1.2, 1.5, 1.8, 2.1, 2.4, 2.7, 3}, and $\overline{\varphi}_i$ is set to {0°,10°,20°,30°,40°,50°,60°, 70°,80°,90°}. For kernel sizes 21 and 31, the range of $L_i$ is expanded, and $\overline{\varphi}_i$ is uniformly distributed between 0° and 90°. The detailed parameter settings can be found in TABLE III.

For feature extraction using deep CNN, training is conducted in cycles of 50 epochs. The initial learning rate of 0.001 is employed for loss optimization with the batch size of 64. The weight $\lambda$ for constraining the open-set risk using the counterpoints is set to 0.1.

## C. Experiments for Closed-Set Recognition

The aim of this experiment is to analyze the proposed method's test accuracy under closed-set conditions and to more intuitively showcase its performance advantages. Due to the relatively standard characteristics of the MSTAR dataset and the attainment of up to 99% accuracy on some common deep learning networks, it's challenging to effectively verify the effectiveness of the study method. Therefore, this chapter devised experiments focusing on limited training samples. In the MSTAR dataset, each object is provided with a full 360° observation angle. Therefore, decreasing the proportion of samples is equivalent to cutting off certain viewing angles of the object, allowing an analysis of the method's robustness against target rotation. The experiment is conducted with training sample sizes of 20, 40, 60, and 80. Here, 'Base' represents the original randomly initialized convolutional kernel, while ASC signifies the kernel designed based on ASC. As shown in TABLE IV, compared to the original convolutional kernel, it's evident from the training process that the ASC-designed kernel achieved significant improvements across all training sample sizes, with respective accuracy enhancements of 9.41%, 1.78%, 4.46%, and 0.66%.

Fig.6 depicts line graphs illustrating the convergence process of models trained with the original convolutional kernel and the designed convolutional kernel under various training sample sizes. As observed from the figure, the model employing the tailored convolutional kernel (ASC) demonstrates a quicker convergence speed and superior accuracy in comparison to the initial model. A preliminary inference suggests that these improvements in results stem from the fact that the ASC-based convolutional kernel model can extract certain features from SAR images earlier and more comprehensively than the original model, particularly nonlinear features that the original convolutional kernel fails to capture. For a detailed comparison and analysis of the visualization effects between the two types of convolutional kernels and their corresponding convolutional layers, please refer to Sections F and G.

TABLE III CONFIGURATION PARAMETERS OF ASC CONVOLUTION KERNELS OF DIFFERENT SIZES

| ASC Kernel Size | | | | | | | Parameters | | | | | | | | | |
|---|---|---|---|---|---|---|---|---|---|---|---|---|---|---|---|---|
| ASC-11 | $L_i$(m) | 0.3 | 0.6 | 0.9 | 1.2 | 1.5 | 1.8 | 2.1 | 2.4 | 2.7 | 3.0 | - | - | - | - | - |
| | $\overline{\varphi}_i$(°) | 0 | 10 | 20 | 30 | 40 | 50 | 60 | 70 | 80 | 90 | - | - | - | - | - |
| ASC-21 | $L_i$(m) | 0.3 | 0.6 | 0.9 | 1.2 | 1.5 | 1.8 | 2.1 | 2.4 | 2.7 | 3.0 | 3.3 | 3.6 | 3.9 | 4.2 | 4.5 | 4.8 |
| | $\overline{\varphi}_i$(°) | 0 | 4.5 | 9 | 13.5 | 18 | 22.5 | 27 | 31.5 | 36 | 40.5 | 45 | 49.5 | 54 | 58.5 | 63 | 67.5 |
| ASC-31 | $L_i$(m) | 0.3 | 0.6 | 0.9 | 1.2 | 1.5 | 1.8 | 2.1 | 2.4 | 2.7 | 3.0 | 3.3 | 3.6 | 3.9 | 4.2 | 4.5 | 4.8 |
| | $\overline{\varphi}_i$(°) | 0 | 3 | 6 | 9 | 12 | 15 | 18 | 21 | 24 | 27 | 30 | 33 | 36 | 39 | 42 | 45 |

| ASC Kernel Size | | | | | | | Parameters | | | | | | | | | |
|---|---|---|---|---|---|---|---|---|---|---|---|---|---|---|---|---|
| ASC-11 | $L_i$(m) | - | - | - | - | - | - | - | - | - | - | - | - | - | - | - |
| | $\overline{\varphi}_i$(°) | - | - | - | - | - | - | - | - | - | - | - | - | - | - | - |
| ASC-21 | $L_i$(m) | 5.1 | 5.4 | 5.7 | 6.0 | 6.3 | - | - | - | - | - | - | - | - | - | - |
| | $\overline{\varphi}_i$(°) | 72 | 76.5 | 81 | 85.5 | 90 | - | - | - | - | - | - | - | - | - | - |
| ASC-31 | $L_i$(m) | 5.1 | 5.4 | 5.7 | 6.0 | 6.3 | 6.6 | 6.9 | 7.2 | 7.5 | 7.8 | 8.1 | 8.4 | 8.7 | 9.0 | 9.3 | - |
| | $\overline{\varphi}_i$(°) | 48 | 51 | 54 | 57 | 60 | 63 | 66 | 69 | 72 | 75 | 78 | 81 | 84 | 87 | 90 | - |

TABLE IV COMPARATIVE EXPERIMENTS OF ADDING ASC MODULE UNDER DIFFERENT SAMPLE SIZES

| Methods | Samples | | | |
|---|---|---|---|---|
| | SOC-20 | SOC-40 | SOC-60 | SOC-80 |
| | Accuracy (%) | | | |
| Base | 53.27 | 70.30 | 81.60 | 87.21 |
| ASC | **62.68** | **72.08** | **86.06** | **87.87** |

## D. Experiments for Open-Set Recognition

The experiments involve evaluating the algorithm's ability to identify and reject unknown samples during the testing phase. Additionally, open-set classification not only emphasizes the classifier's success in rejecting unknown samples but also underscores its accuracy in classifying known categories. Following prior research [43], the experiments select 2S1, BRDM2, BTR60, D7, T62, ZIL131, ZSU23/4 from the MSTAR database as the training set, reserving the remaining three classes as unknown targets. This setup aids in assessing the proposed method's performance when confronted with unknown targets. The experiments employed data from seven known classes to train the model, while the test set comprised both known and unknown categories, facilitating inference and final performance evaluation. Each experimental group underwent three repetitions to ensure the reliability and consistency of the outcomes.

This section encompasses various mainstream open-set methods for comparison, including MLS[14], CBC[43], and OpenMax[44]. These methods possess distinct characteristics and strategies in the field of open-set recognition.

The four indicators presented in TABLE V are utilized to evaluate the performance of our proposed algorithm. As evident from the TABLE V, our method demonstrates exceptional performance, surpassing other methods in all four assessment metrics, thereby highlighting its significant advantages. Moreover, to comprehensively evaluate the performance variations of various methods under different degrees of openness, the known class count is gradually increased from 3 to 7 in increments of 1, and Equation (25) is used to compute the corresponding openness. Fig. 7 illustrates the F1 scores and accuracies under different degrees of openness. Notably, the proposed method exhibits relatively stable F1 scores and accuracies, highlighting its robustness. In contrast, other methods showed significant fluctuations under varying degrees of openness. This stability represents a crucial advantage of the proposed method, particularly suitable for practical applications in diverse open environments. Through in-depth analysis of the results and curve trends depicted in Fig. 7, this study derives valuable conclusions that contribute to a better comprehension of open-set classification performance:

(1) When the openness is 0, it corresponds to the closed-set performance of each algorithm, indicating the model's performance solely within known category scenarios. As unknown categories are encountered, all methods experience a decline in performance. Yet, the extent of this decrement serves as a barometer of a model's proficiency in open environment. Typically, a more modest decline in performance indicates greater stability of the model in dealing with unknown categories.

(2) The open-set performance of classification methods is positively correlated with their closed-set performance. This suggests that methods exhibiting excellent open-set performance tend to also demonstrate better closed-set performance, indicating a positive correlation between the algorithm's ability to identify unknown categories and its capacity in feature extraction.

TABLE V QUANTITATIVE EVALUATION INDICATORS OF DIFFERENT METHODS ON THE MSTAR DATA SET

| Method | Precision | Recall | F1 score | Accuracy |
| --- | --- | --- | --- | --- |
| MLS | 84.72 | 87.78 | 86.32 | 86.65 |
| OpenMax | 85.73 | 72.65 | 74.14 | 73.73 |
| CBC | 89.16 | 85.82 | 87.05 | 86.75 |
| Ours | **91.08** | **88.75** | **89.64** | **88.98** |

TABLE VI THE IMPACT OF DIFFERENT SIZE CONVOLUTION KERNELS AND ASC

| Size | Samples | | | | | | | |
| --- | --- | --- | --- | --- | --- | --- | --- | --- |
| | SOC-20 | | SOC-40 | | SOC-60 | | SOC-80 | |
| | Base | ASC | Base | ASC | Base | ASC | Base | ASC |
| 31 | 53.27 | 62.68 | 70.30 | **72.08** | 81.60 | **86.06** | 87.21 | **87.87** |
| 21 | 49.93 | **62.76** | 68.86 | 71.75 | 77.56 | 82.14 | 83.91 | 85.36 |
| 11 | 48.74 | 61.52 | 67.71 | 69.11 | 79.29 | 79.79 | 82.06 | 83.58 |

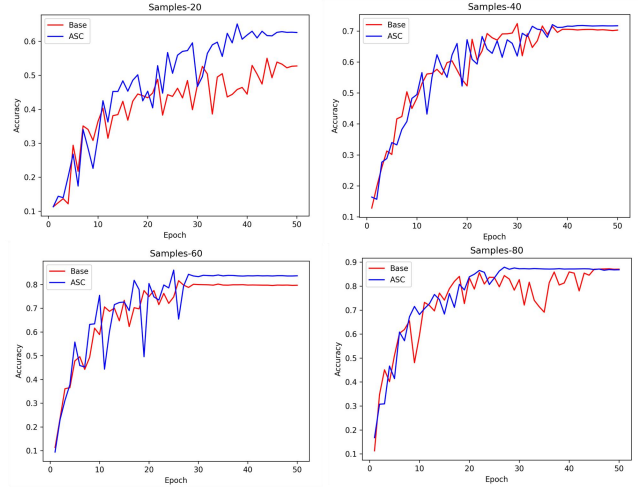

Fig. 6 Comparison of the convergence process of Base and ASC modulated convolution kernels

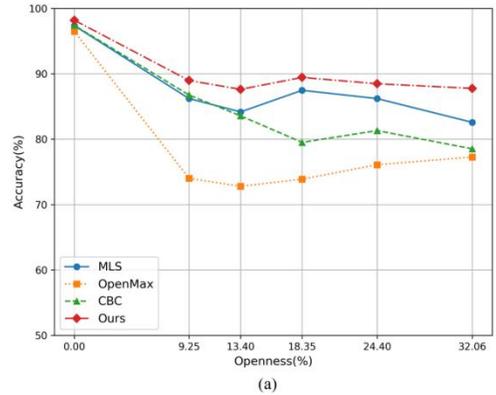

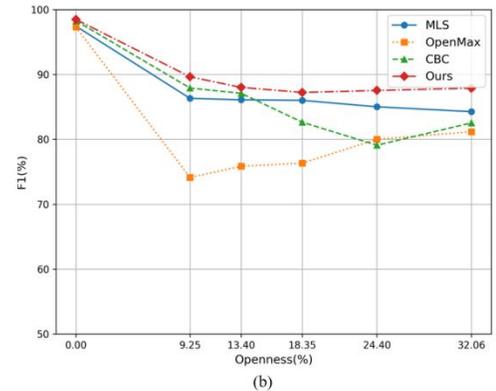

Fig. 7 Results of different openness on the MSTAR dataset (a) Accuracy (b) F1 score

## E. Ablation Study

### 1) The effectiveness of large-scale kernel and ASC

According to the results in TABLE VI, under the Base algorithm, as the convolutional kernel size increases from 11 to 31, the accuracy exhibits varying degrees of improvement: 4.53%, 2.59%, 2.31%, and 5.15% respectively. The potential rationale behind this trend could be attributed to the introduction of large kernels, which introduce more shape biases into the network. In essence, ImageNet images can be correctly classified based on texture or shape. However, human perception primarily relies on shape cues rather than texture for object recognition. Therefore, models with stronger shape biases may better generalize to the target task. In the context of ASC augmentation, the accuracy relative to the Base algorithm increases by 13.94%, 4.37%, 6.77%, and 5.81% respectively. The underlying cause is attributed to the designed large-sized ASC kernel, which, compared to traditional convolutional kernels, provides a larger receptive field. Moreover, it demonstrates higher focus in extracting certain intrinsic non-linear features and specific scattering characteristics in SAR images, as evident in the visualizations in Fig. 8 and Fig. 10.

### 2) The Visual Analysis of feature map

To exhibit the efficacy of the study visually, we created visualizations of the output feature maps from both the initial and fifth convolutional layers for both approaches. As depicted in Fig.8, the first convolutional layer preserves nearly all the original image details, while deeper layers abstract the extracted features. This signifies a reduction in visual content information with increasing network depth, accompanied by an increase in distinctive category-related features. A distinct observation from Fig.8(b)(d) illustrates that the designed model's output features are more focused on the target, unlike the original model which might be affected by background noise interference.

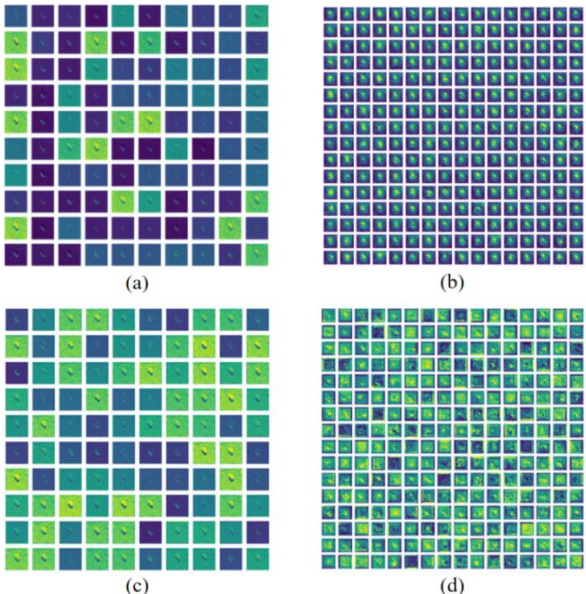

Fig. 8 Visualization of the output feature map (a) (b) the output feature maps of the first and fifth convolutional layers based on the ASC model respectively; (c) (d) the output feature maps based on the first and fifth convolutional layers of the original model respectively.

### 3) The effectiveness of RPL

As indicated in TABLE VII, with the help of RPL, compared with the basic method, the algorithm achieves significant improvements in overall accuracy and F1 score, which is mainly attributed to the correct classification of unknown categories. Further, to demonstrate the recognition advantage between known and unknown classes more clearly, we perform feature space visualization on their penultimate layer. As depicted in Fig.9(a), significant overlap exists in feature distribution between unknown and two known classes (2S1, BTR60), resulting in erroneous classification of unknown samples as known ones due to overlapping feature distributions. At the same time, a distinct delineation is evident in the feature space between the unknown class and the seven known classes, with no overlap observed amongst the latter in Fig.9(b). Specifically, the utilization of reciprocal point learning has facilitated the creation of a well-defined feature space, within which the known classes are effectively marginalized, thereby enhancing the identifiability of the unknown class.

TABLE VII THE EFFECTIVENESS OF RPL

| Method | Precision | Recall | F1score | Accuracy |
|---|---|---|---|---|
| Base | 74.73 | 70.65 | 72.46 | 70.13 |
| ASC-RPL | 91.08 | 88.75 | 89.64 | 88.98 |

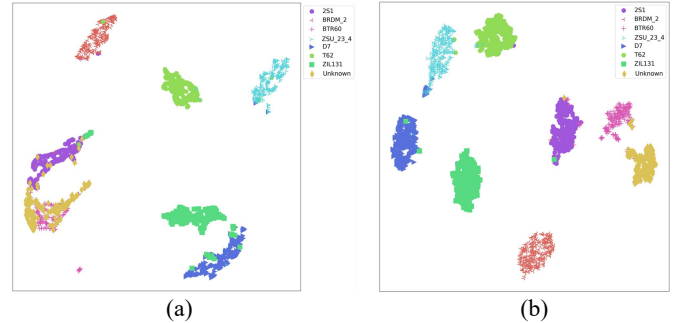

Fig. 9 Feature space visualization. (a) Original feature space (b) Feature space constructed by the proposed method.

## F. Visual Analysis of Convolution Kernel Base ASC

Fig. 10 displays the amplitude images of convolutional kernels designed based on the ASC model. These kernels have a size of 11×11, with a quantity of 100 and a single channel. The ASC-designed convolutional kernels exhibit good orthogonality in their kernel weights due to different combinations of $\{L_p, \varphi_{0p}\}$. Compared to regular convolutional kernels, networks modulated by ASC achieve higher average precision. This could be attributed to SAR responses originate from microwaves rather than visible light. As is well-known, in the field of optical image, the continuous linear structures are derived from the first convolutional layer[20]. However, in SAR images, the complete linear structure is only visible at a zero azimuth angle (as seen in Fig. 10). When the scattering center points in other directions, the linear structure crosses distance bins and breaks. This might be the primary reason why ASC-designed convolutional kernels outperform regular ones in SAR target recognition. Fig. 11 presents visualizations of convolutional kernels with a size of 21x21, totaling 441 kernels. Compared to the 11x11 kernels, the 21x21 kernels can extract more abundant scattering features.

*G. The Heatmap Visualization*

This present paper employs visualization techniques to analyze the decision-making process of the model. Specifically, The Grad-CAM is utilized for visualizing and interpreting the decision mechanism of CNNs. Grad-CAM generates heatmaps representing important regions for category determination by analyzing feature maps and the gradient backpropagation information for a specific target class. GradCAM++ enhances Grad-CAM by introducing regularization terms to improve the quality of visualization. XGradCAM extends GradCAM and GradCAM++ by incorporating additional regularization terms, further enhancing the visualization effects. EigenCAMs utilize eigenvalue decomposition to generate heatmaps based on the feature maps of convolutional layers, aiming to improve interpretability. These methods collectively contribute to explaining the decision-making process of CNNs, enhancing the understanding of model predictions, and providing robust support for the interpretability of deep learning models.

As shown in Fig. 12, this study employed four visualization techniques to generate heatmaps for two models. It can be observed that, the target region exhibits broader activation in (II), concentrating the heatmap more significantly on the target and its components, thereby emphasizing the focal regions of the model. In contrast, the background region in (I) is activated by mistake. This indicates that convolutional kernels, crafted utilizing ASC information, are adept at accurately perceiving the target area while diminishing the effects of background noise on recognition, thereby substantially enhancing the model's robustness.

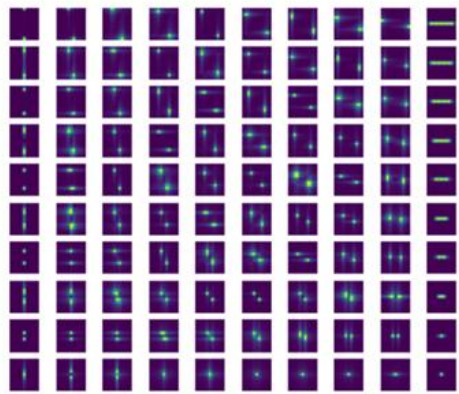

Fig. 10 Visualization of convolution kernel with size 11x11

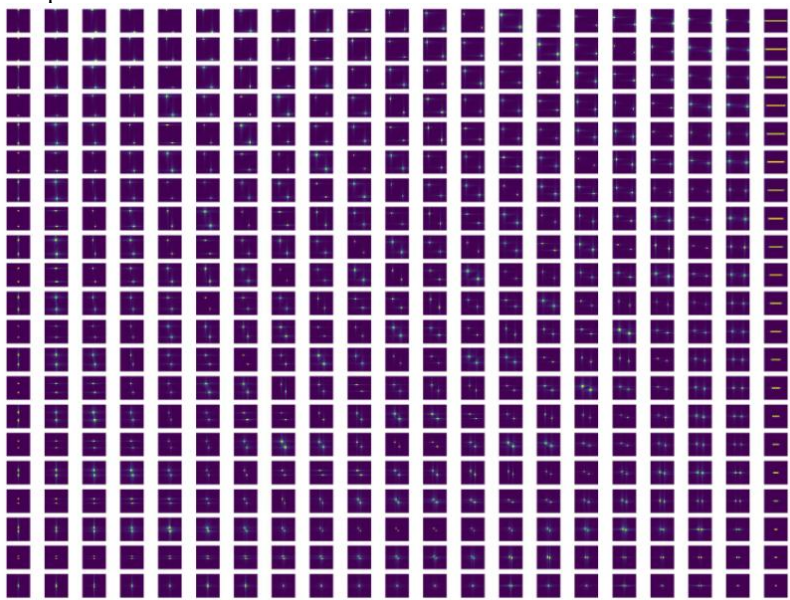

Fig. 11 Visualization of convolution kernel with size 21x21

V. CONCLUSION

The typical evaluation processes of the popular CNN algorithm are limited to closed environments, failing to adequately meet the practical demands and variations of open environments. Common closed-set detection methods, constrained by limited training samples, often struggle to effectively handle unknown categories in open environments. Therefore, a novel approach cooperated with scattering mechanism is designed for optimizing the performance of SAR image recognition in open sets.

The motivation behind this study encompasses two crucial aspects: Firstly, a feature learning framework based on reciprocal point learning is established to simulate extraneous spaces, thereby mitigating risks in open environments; Secondly, the learned feature space demonstrates the capability to augment differences between known and unknown categories, reducing overlap between their respective distributions; Thirdly, a large-scale convolutional kernel network integrating the attribute scattering centers is designed to better model SAR image scattering features, capturing richer spatial semantic information. This enables the backbone network to acquire robust feature representations for identifying both known and unknown categories within a refined feature space. Comprehensive experiments validate the effectiveness of the proposed framework.

Traditional closed-set recognition primarily focuses on distributional changes in the known samples, thus failing to adapt well to changes in the environment. Based on this, models must extend known categories into a more extensive open space. In the future, we aim to expand this research into open-set detection, encompassing the identification of unknown samples as well as the task of detecting and classifying them as new categories.

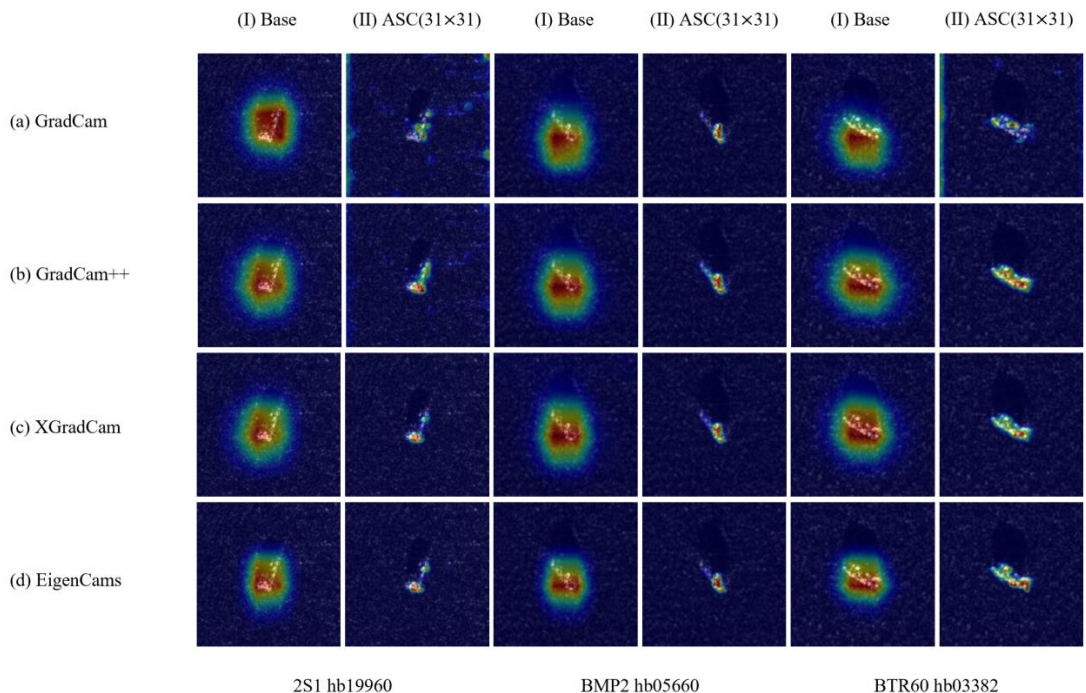

Fig. 12 Heatmap of test images. (I) Base method (II) The proposed method.